# Backpropagation and fuzzy algorithm Modelling to Resolve Blood Supply Chain Issues in the Covid-19 Pandemic

Aan erlansari[1], Rusdi Efendi[2], Funny Farady C[3], Andang Wijanarko[4], Reza Herliansyah[5], Boko Susilo[6]

Faculty of Engineering, University of Bengkulu

aan_erlanshari@unib.ac.id, r_efendi@unib.ac.id, ffarady@unib.ac.id, andang@unib.ac.id, reza@gmail.com, bsusilo@unib.ac.id

**Abstract**

Bloodstock shortages and its uncertain demand has become a major problem for all countries worldwide. Therefore, this study aims to provide solution to the issues of blood distribution during the Covid-19 Pandemic at Bengkulu, Indonesia. The Backpropagation algorithm was used to improve the possibility of discovering available and potential donors. Furthermore, the distances, age, and length of donation were measured to obtain the right person to donate blood when it needed. The Backpropagation uses three input layers to classify eligible donors, namely age, body, weight, and bias. In addition, the system through its query automatically counts the variables via the Fuzzy Tahani and simultaneously access the vast database.

## I. PRELIMINARY

Blood is one of the important tissues in the human body that has lots of specific functions. [1][2]. An example include transporting oxygen (O2) and nutrients and releasing toxins within the body. Furthermore, this explains the theory which states that the human body contains around 7 to 8% of blood.

This tissue is made up of three components, namely erythrocytes, leukocytes, thrombocytes, and plasma[3]. These components has a specific function to maintain human health. Furthermore, it can be transfused from one person circulatory system to another due to certain medical conditions such as trauma, surgery and shock. There are eight ABO blood types, and it is preferable to transfuse blood between patients of the same blood match in order to prevent the immune system from attacking the tranfused red blood cells.

The blood supply chain (BSC) manages the flow of blood products from donors to patients through five echelons, namely donors, mobile collection sites (CSS), blood centers (BCs), demand nodes, and patients. The demand nodes include hospitals, clinics, or other transfusion points. Furthermore, mobile CSS, BCs, and demand nodes need to be coordinated in other to perform the six main processes associated with blood donation. They include collection, testing, component processing, storage, distribution, and transfusion[4].

The shortage of bloodstock and its uncertain demand has become a significant problem for all countries worldwide. A sufficient population of donors need to be available in order meet the needs for transfusion within a reasonable period of time.. Inadequate bloodstock during the Covid-19 pandemic was mentioned many times at Bengkulu Province. Furthermore, between 2019 to 2020[5], the blood unit's realization was around 16.000 blood packs, and it was still far enough from the target of 22.000 blood units[6]. Moreover, the Red Cross faced a problematic situation in order to arrange a massive donation from a donor mobile. It is believed that when this current situation extends for a long period of time, Bengkulu would face a serious problem with no blood supply in their blood bank.

The Indonesian Red Cross chapter at Bengkulu began to collect donors data few years ago in order to manage their activities, schedule, and blood type. However, the data collected was not able to handle inadequate bloodstock during the Covid-19 Pandemic. Figure 1 [7] displays the current distribution scheme, whereby donors (suppliers) periodically come to the Indonesian Red Cross center (PMI) to donate blood or via the Mobile Collection sites (CSS). PMI is the legal organization processing the blood from the donor into final products such as whole blood, plasma, thrombocyte, and others. Furthermore, PMI distributes the final products to hospitals or other health facilities.

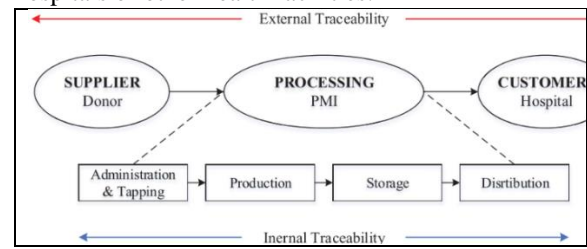

Fig 1 Blood distribution model

This problem was solved by developing a blood supply management system using Backpropagation algorithm and BSC evaluation using Fuzzy Tahani.

## II. BACKPROPAGATION AND FUZZY ALGORITHM

### A. Backpropagation

This has become the most popular method of training neural networks due to the underlying simplicity and relative power of the algorithm.

Backpropagation[8] is a learning algorithm used in reducing the error rate by adjusting the weight based on the difference in the output and the desired target. It is also defined as a Multilayer training algorithm because of its three layers, namely input, hidden, and output. Backpropagation involves developing a single-layer network with two layers, namely the input and output. Fig one shows the schematic diagram of a two-layered feed-forward network employing full connectivity between adjacent layers.

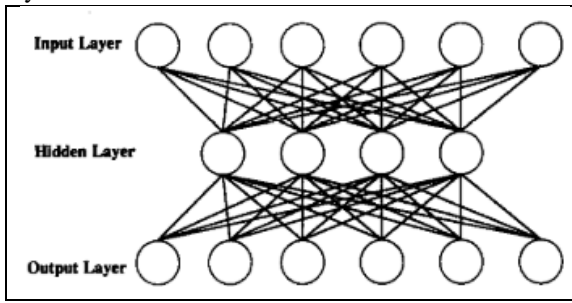

Fig. 2 Backpropagation network layer Full connection

Based on the to figure above, the 'input layer' performs no processing on its inputs and mainly distribute them to the first processing layer. While the hidden layer, receives no input and produces no output. Finally, the 'output layer' produces the output results of the network for the user. The number of input nodes is fixed by the number of input variables provided for the task. While the number of output nodes is fixed by the number of values that are desired.

### B. Fuzzy Tahani

Fuzzy logic is the best way of mapping an input into an output area for any complicated issue. Its basic concept is to perform a calculation on input variables based on its disguised value.

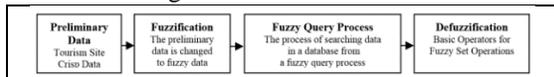

Fig 3. Fuzzification process

The figure above shows all data related to parameters and alternatives from blood donor attraction and preparedness. Furthermore, the membership function is a curve that shows the use of data and has an interval between 0 and 1 in its membership value. It is also present in a fuzzy system which has a combination membership degree between the left shoulder, triangle, and right shoulder curves. Each function's domain starts from 0 to ∞ (infinite), in order for the procedure's domain to become more flexible.

Fuzzification is the conversion of crisp values to that of fuzzy. Furthermore, a fuzzy Inference System (FIS) is in charge of making conclusions from a set of rules. Therefore,, the FIS results in this study would be used to determine the value of recommendations from age, distance, and time attractions. Query fuzzification is assumed to be a conventional query database management system that creates and implements a basic system of fuzzy query logic.

## III. RESULT AND ANALYSIS

In order to overcome the issues explained in section 1, The backpropagation algorithm that evaluates donation possibility was used in solving the issues in section 1. The donor were initialized $O_1$ and $O_2$ for possible and impossible donors. Furthermore, two types of input combined with cross-validation from the table below were used.

Table 1. cross-validation

| Attribute | Value |
| --- | --- |
| Fold number | 10 |
| Training cycle | 100 |
| Learning rate | 0.001 |
| Hidden layer | 1 |
| Neuron | 3 |
| Momentum | 0.9 |
| Error epsilon | 0.001 |
| Activation | Sigmoid (0-1) |

The results from the above table is described in the figure below..

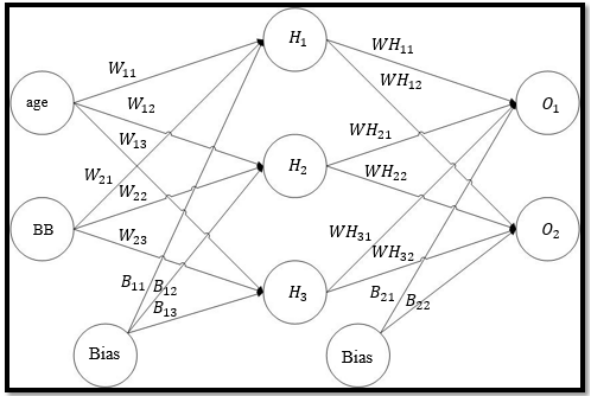

Fig 4. Backpropagation evaluation

Using this equation:
$$h_1 = i_1 \times wi_{11} + i_2 \times wi_{21} + b_1 \quad (1)$$

Initiation from Figure above we have:

Input: $Usia = i_1 = 0{,}827; BB = i_2 = 1$

Bias:
$B_{11} = -2.557; B_{12} = -2.436; B_{13} = -1.608$
$B_{21} = -3.690; B_{22} = 3.7$

Weight:
$W_{11} = 2.646; W_{12} = 2.530; W_{13} = 1.785; W_{21} = 2.581; W_{22} = 2.462; W_{23} = 1.676; WH_{11} = 3.360; WH_{12} = -3.327; WH_{21} = 3.093; WH_{22} = -3.066; WH_{31} = 1.580; WH_{32} = -1.658;$

And,
$H_1 = (0{,}827 \times 2.646) + (1 \times 2.581) - 2.557$

$$out\ h_1 = \frac{1}{1 + e^{-h_1}} = \frac{1}{1 + e^{-0.3775}} = 0.593269992$$

$out\ h_2 = 0.596884378$

$o_1 = out\ h_1 \times wo_{11} + out\ h_2 \times wo_{21} + b_2$

$o_1 = 0.3775 \times 0.40 + 0.596884378 \times 0.45 + 0.16 = 1.105905967$

$$out\ o_1 = \frac{1}{1+e^{-h_1}} = \frac{1}{1+e^{-1.105905967}} = 0.75136507$$

$out\ o_2 = 0.772928465$

$O_1 = 3{,}4606$

$$Out(O_1) = \frac{1}{(1+e^{-O_1})} = \frac{1}{(1+e^{-3.4604})} = 0.9663$$

$O_2 = H_1 \times WH_{12} + H_2 \times WH_{22} + H_3 \times WH_{32} + B_{22}$

$O_2 = 0.9012 \times -3.327 + 0{,}8926 \times -3.066 + 0.8239 \times -1.658 + 3.7$

$O_2 = -3.517$

$$Out(O_2) = \frac{1}{(1+e^{-O_2})} = \frac{1}{(1+e^{+3.517})} = 0.0338$$

Based on the calculation above, the result of confident $O_1$ was 0.9663 and confident $O_2 = 0.0338$. In Tahani databases, initially, a fuzzy set formed with its membership function. In order to access the available donor, several categories were constructed, namely ages, distance, and time.

The database and its sturructure used in this study was Tahani, and relational. Furthermore, the selected data was processed using the Fuzzy Tahani method with the parameters desired by the donor. Each variable fuzzy membership function used a left shoulder, triangle, and right shoulder curves for three fuzzy sets.

A. Age

Figure 4 presents the membership parameter in the domain of functions 1 (Age) which is divided into three parts, namely Muda (young) which is between 0 to 33, Paruh baya (middle age) which is between 17 to 60, and Tua (elderly) which is between 33 to 60.

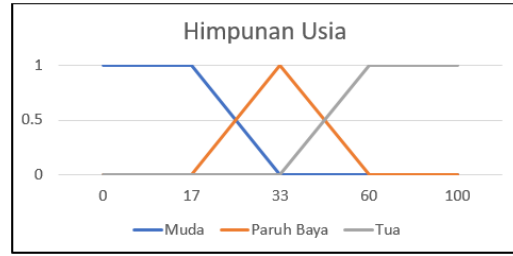

Fig 5. Set of age

The age parameter stated that the higher value was 1 when the age was at 60.

Therefore, the calculation from the existing formula for age parameter is presented below.

$$\mu_{Tua}[X] = \begin{cases} 0; & x \leq a \\ \frac{x-a}{b-a}; & a \leq x \leq b \\ 1; & x \geq b \end{cases} \quad (2)$$

$$\mu_{Tua}[Usia] = \begin{cases} 0; & x \leq 33 \\ \frac{x-33}{60-33}; & 33 \leq x \leq 60 \\ 1; & x \geq 60 \end{cases}$$

Table 2. age data set

| Donor | age (y) | Derajat Keanggotaan | | |
|---|---|---|---|---|
| | | Muda | Baya | Tua |
| Person 1 | 38 | 0 | 0.815 | 0.182 |
| Person 2 | 42 | 0 | 0.667 | 0.333 |
| Person 3 | 37 | 0 | 0.852 | 0.148 |

### B. Distance

Figure 5 present the membership parameter in the domain of functions 2 (distance) which is divided into three parts, namely jauh (far) 0 to 10000m, dekat (near) 5000m, agak jauh (a bit far)

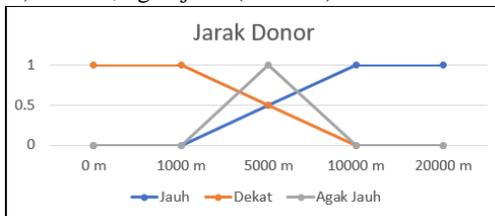

Fig 6. Set of distance

The distance parameter stated the higher value was 1 when the distance was at 10000

The fuzzy Tahani equation used in calculating the prior distance is shown below.

$$\mu_{Jauh}[X] = \begin{cases} 0; & x \leq 1000 \\ \frac{x - 1000}{10000 - 1000}; & 1000 \leq x \leq 10000 \\ 1; & x \geq 10000 \end{cases}$$

Furthermore, the result of the distance parameter calculation is shown in table 3.

Table 3. distance data set

| Name | Distance | Degree membership | | |
|---|---|---|---|---|
| | | Near | A bit Far | Far |
| Person 1 | 1302 | 0.966 | 0.076 | 0.034 |
| Person 2 | 4835 | 0.574 | 0.959 | 0.433 |
| Person 3 | 8109 | 0.210 | 0.378 | 0.7899 |

### C. Donor Time

The membership parameter in the domain of function three (donor time) is presented in Figure 6, and is divided into two parts, namely *lama* (old) 0-90 days, *baru* (recent time), agak lama (short-term)

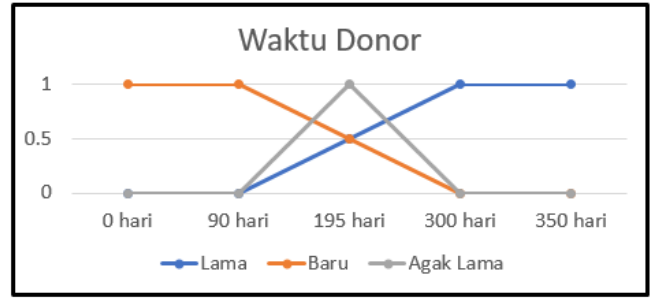

Figure 7. set of donor time

The distance parameter stated the higher value was 1 when the donor time was at 300 days or less than 90

Table 4. time data set

| Name | Time | Degree membership | | |
|---|---|---|---|---|
| | | Recent time | Short-term | Old |
| Person 1 | 270 | 0.143 | 0.286 | 0.857 |
| Person 2 | 158 | 0.676 | 0.648 | 0.324 |
| Person 3 | 320 | 0 | 0 | 1 |

All the data set formed and calculated are shown in table 2, 3, and 4. Furthermore, Tahani was able to perform the evaluation process using the query from the database:

*Select \* From Data Pendonor Where (Jarak = "Dekat") And (Usia = "Baya") And (Waktu Donor = "Lama")*

Furthermore, the results in the table below were obtained using these parameters and query.

Table 5. Evaluation using Tahani and Backpropagation

| Nama Pendonor | Dekat | Baya | Lama | Prioritas |
|---|---|---|---|---|
| Erikson | 0.966 | 0.815 | 0.857 | 0.857 |
| Deddy dinpansyah | 0.574 | 0.667 | 0.324 | 0.324 |
| Yetti Sukmawati | 0.210 | 0.852 | 1 | 0.210 |

## IV. Conclusion

This study described the whole process of implementing Backpropagation in order to specify eligible donors. Based on Backpropagation's three input layers, the donor was identified by its blood type, age, and weight. The selection criteria was for ages between 17 to 60 years, and body weight above 40 kg. Outside these category donors were automatically eliminated by the system.

The Fuzzy Tahani algorithm classified the potential donors based on age, distance, and time to last donation. Furthermore, in order to support mobility of blood supply chain management this algorithm was embedded directly into the system using a specific query to access the database. The effectiveness of using the Backpropagation and Fuzzy Tahani produced 99.5% accuracy when selecting eligible and potential donors.